\documentclass[10pt,twocolumn,letterpaper]{article}

\newcommand\blfootnote[1]{%
  \begingroup
  \renewcommand\thefootnote{}\footnote{#1}%
  \addtocounter{footnote}{-1}%
  \endgroup
}

\usepackage{cvpr}              

\usepackage{longtable}
\definecolor{cvprblue}{rgb}{0.21,0.49,0.74}
\usepackage[pagebackref,breaklinks,colorlinks,allcolors=cvprblue]{hyperref}

\def\paperID{17487} 
\def\confName{CVPR}
\def\confYear{2025}

\title{Towards Predicting the Success of Transfer-based Attacks by Quantifying Shared Feature Representations}

\author{Ashley S. Dale\textsuperscript{\dag1}*
\and
Mei Qiu\textsuperscript{\rm1}
\and
Foo Bin Che\textsuperscript{\rm1}
\and
Thomas Bsaibes\textsuperscript{\rm 2}
\and
Lauren Christopher\textsuperscript{\rm 1}
\hspace{1cm}
Paul Salama\textsuperscript{\rm1}*\\
\\
\textsuperscript{\rm 1}Department of Electrical and Computer Engineering, Purdue University\\
\textsuperscript{\rm 2}Department of Physics, Indiana University Indianapolis
}

\begin{document}
\maketitle
\begin{abstract}
Much effort has been made to explain and improve the success of transfer-based attacks (TBA) on black-box computer vision models. This work provides the first attempt at a priori prediction of attack success by identifying the presence of vulnerable features within target models. Recent work by Chen and Liu (2024) proposed the manifold attack model, a unifying framework proposing that successful TBA exist in a common manifold space. Our work experimentally tests the common manifold space hypothesis by a new methodology: first, projecting feature vectors from surrogate and target feature extractors trained on ImageNet onto the same low-dimensional manifold; second, quantifying any observed structure similarities on the manifold; and finally, by relating these observed similarities to the success of the TBA. We find that shared feature representation moderately correlates with increased success of TBA $\left(\rho = 0.56\right)$. This method may be used to predict whether an attack will transfer without information of the model weights, training, architecture or details of the attack. The results confirm the presence of shared feature representations between two feature extractors of different sizes and complexities, and demonstrate the utility of datasets from different target domains as test signals for interpreting black-box feature representations.
\end{abstract}    
\blfootnote{\dag Work completed at Purdue University, now at University of Toronto}
\blfootnote{*Corresponding Author Emails: ashley.dale@utoronto.ca, psalama@purdue.edu}
\section{Introduction}
\label{sec:intro}
A transfer-based attack (TBA) uses a surrogate model to generate attacks,  which are then transferred to a target model~\cite{papernot2016transferability}.  These attacks are leveraged through the transfer of gradients from the surrogate to the target~\cite{sen2024fgsm}, or through query methods which rely on responses from the target to inputs fine-tuned on the surrogate~\cite{liang2022parallel}. The primary benefit of this approach is the ability to target black-box models~\cite{bhagoji2018practical}, as TBA eliminates the need for information about the target model's gradients and/or weights. Much effort has been made to improve the success of TBA, with continuous contributions of new methodologies~\cite{cheng2019improving, wu2023black, li2023sibling, fang2022learning, chen2024content} and evaluation approaches~\cite{li2024towards}.  

\begin{figure*}[!ht]
    \centering
    \includegraphics[width=\textwidth]{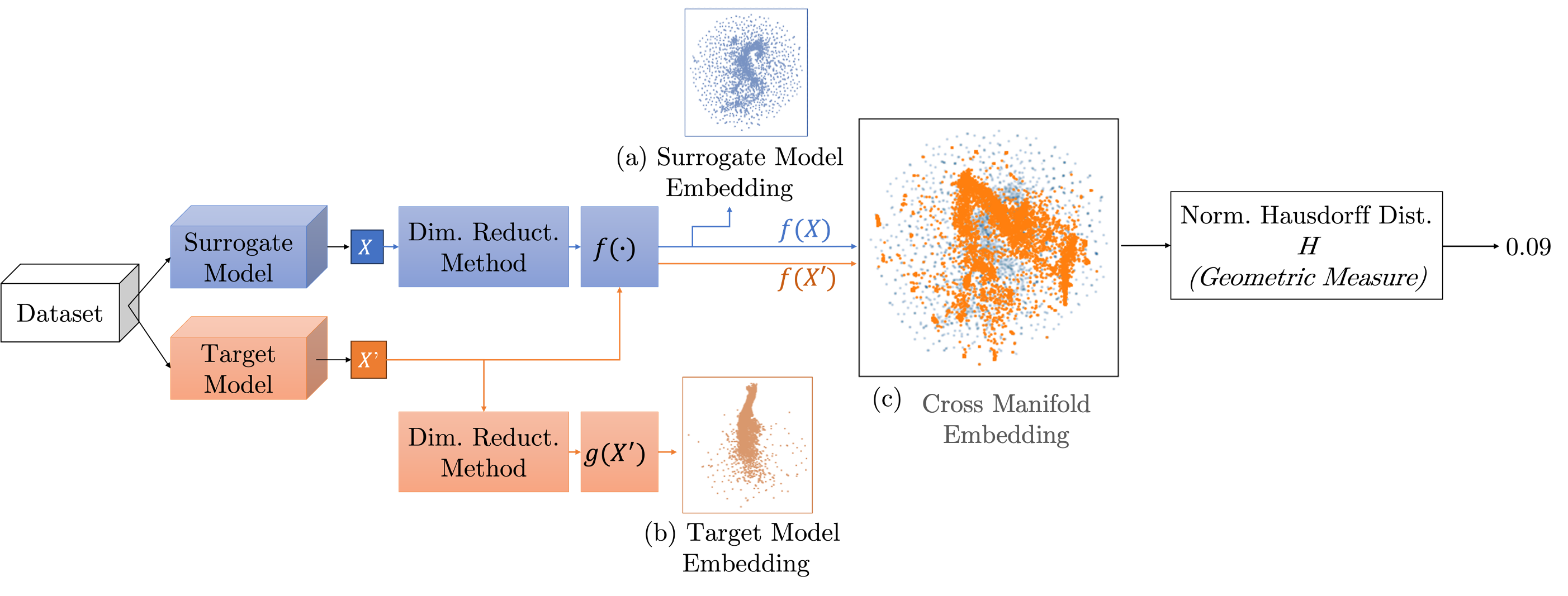}
    \caption{Summary of methodology used to identify shared low-dimensional manifolds. A surrogate model and target model are chosen, then used to create feature vector representations of the same dataset.  Dimensionality reduction via the UMAP algorithm enables the cross projection of feature vectors from the target model onto the manifold of the surrogate model. Results are shown for the SI-Score~\cite{djolonga2020robustness} dataset, with ResNetv2 as the surrogate model and MobileNetv3 as the target model.}
    \label{fig:results_siscore}
\end{figure*}

This work is the first to ask \textit{\textbf{``is it possible to predict the success of a transfer-based attack on a black box model without any a priori knowledge of the attack?"}} Answering this question is challenging, but valuable as it allows the identification of potential model vulnerabilities.  At this time, the only other work (to our knowledge) seeking to predict the success of black box attacks was carried out by Puccetti et al.~\cite{puccetti2023attacks} and also requires information about the nature of the attack.  We are the first (to our knowledge) to formulate the question with the stated constraints on a priori information, and the first to answer in the affirmative. Our attempt to answer this question is presented through the form of a new methodology, as visualized in Figure~\ref{fig:results_siscore} and applied to a toy example.

The results presented here challenge the TBA community to better identify and quantify feature representations which make models vulnerable to transfer-based attacks.  The discussion section presents several suggestions for future work, as the method presented here promises to allow access to new research directions, and ultimately yielded more questions than answers.

\subsubsection*{Main Contributions}
Our main contributions lie in three key areas: 

\textbf{First}, because this work is the first to predict transfer-based attack success using minimal information, we introduce criteria for evaluating methodologies that attempt to accomplish this task. These criteria establish a \textbf{new research direction}, enabling the identification of vulnerable models without knowledge of the attacks or model weights.

\textbf{Second}, we show that the shared feature representations previously proposed to explain the success of transfer-based attacks can be identifiable and quantifiable by analyzing the feature vectors returned by the feature extraction backbones of large CNN models used for various computer vision tasks.  We accomplish this by \textbf{introducing a \textit{cross manifold embedding} methodology} that places constraints on the latent space structure of the models instead of directly inspecting model weights.  The use of a dimensionality-reduction model (e.g. PCA \cite{greenacre2022principal} and UMAP \cite{mcinnes2018umap}) trained on a subset of data samples to project new samples onto the lower-dimensioned manifold is now widely accepted \cite{snyder2021connected, rudar2023decision, fan2024cohortfinder, hofer2022model}.  In previous works using this method, the dataset feature representations are fixed and the dimensionality of the feature set is a known constant. This work is the first to extend the method to samples with divergent feature dimensionalities and feature representations by using a ``cross-projection" method tuned to one of the two feature spaces. A primary result of this work is that cross-projection of the latent feature space from one model onto the latent space of another model is meaningful, even when the dimensions of the two latent spaces are different.  This is discussed further in \S \ref{ssec:shared_feat_rep}.

\textbf{Third}, we use the criteria and  methodology proposed in this work to show that observed similarities in shared feature representation correlate with transfer-based attack success, where increased similarity generally indicates increased success.  The correlation between similar feature representation and transfer-based attack success is quantified using the correlation coefficient and the eigenvalues from Principal Component Analysis decomposition (PCA) \cite{greenacre2022principal} as applied to the joint distribution of transfer-based attack success values and the Hausdorff distance between cross manifold embeddings. This leads to the \textbf{first predictive method for transfer-based attack success without the need for additional information} about model training, weights, architecture, or attack details.

\section{Background}
\label{sec:background}
\subsection{Shared Feature Representation through Cross Manifold Embeddings}
\label{ssec:shared_feat_rep}
To answer the question of predicting transfer-based attack (TBA) success on a black-box model, we propose that identifying shared feature representations between the target and surrogate models is meaningful for evaluating TBA success. The main challenge of this work is developing a valid method which identifies these features in a black-box context \textit{without knowledge of model weights, gradients, training parameters, or architecture}.  We term the aggregate of feature representations produced by the model's feature extraction backbone in vector form as the model's \textit{feature space}, differentiating from the model's latent space as encoded by the weights of the feature extraction backbone, where the weights are inaccessible in a black-box model study.

Our primary hypothesis is that \textit{\textbf{a shared low-dimensional feature space between two models is identifiable using dimensionality reduction methods on samples of the model's feature space.  A shared low-dimensional manifold will have a quantifiable shared dimensionality, while disparate manifolds will be orthogonal.}}  For this work, we seek to determine whether a dataset $D$ represented by sets of feature vectors \{$D_S, D_T$\} extracted from surrogate and target models respectively can be equivalently transformed using a dimensionality reduction algorithm $f(\cdot)$ s.t. the equality of the sets after transformation is $f(X) = f(X')$, $X\in D_S$, $X'\in D_T$; there is no assertion that the embeddings represent the true underlying structure of the dataset. There is also no assertion in this study that the shared low-dimensional space presented maximizes shared semantic meaning across all possible subspaces between the target and surrogate models.  Instead, we assert that the lack of optimization when choosing the shared subspace between models in this work validates the proposed methodology, where (according to the lottery ticket hypothesis~\cite{da2022proving}) finding a pair of ``winning tickets'' in the target and surrogate models has implications regarding the total number of shared winning tickets available: to find any winning ticket suggests that many winning tickets are available to be found, and this explains and motivates the success of a transfer-based attack. 

\subsection{Related Works}

\subsubsection*{Theory of transfer-based attacks}

Various explanations for the success of transfer-based attacks have been presented~\cite{chen2023rethinking, qin2023training, fang2022learning}.  Most recently, the manifold attack model~\cite{chen2024theory} unified previous explanations by formalizing the concept of shared model feature spaces: for a transfer-based attack to be successful, the adversarial examples must exist in a shared lower-dimensional manifold. The experimental results presented in~\cite{chen2024theory} are limited to demonstrations of the proofs; in this work, we extend this main idea to popular feature extraction backbones used for computer vision tasks.

\subsubsection*{Shared lower dimensional manifold assumption}

The strong lottery ticket hypothesis states that multiple lower-dimensional solutions to the same problem can exist within the same model architecture~\cite{da2022proving}.  This motivates techniques such as model weight quantization~\cite{zhang2023lottery}, and model weight pruning and knowledge distillation~\cite{he2023structured}.
The latter two techniques are dimensionality reductions of the model latent space, which naturally extend to methods such as PCA~\cite{greenacre2022principal}, T-SNE~\cite{anowar2021conceptual}, and UMAP~\cite{allaoui2020considerably} performed on samples from high dimensional latent spaces to reduce the dimensionality.  For linearly separable data, PCA is the uncontested standard.  For non-linear spaces, UMAP's original algorithm suggests that distance metrics on UMAP embeddings are meaningful~\cite{mcinnes2018umap}, as it attempted to approximate the original high-dimensional space.  However, this understanding was later revised~\cite{damrich2021umap}, and UMAP can now be interpreted as a refinement of the $k$NN graph~\cite{damrich2021umap}. We choose UMAP for this study, as the data is non-linear and the analysis of the manifold requires only the justified use of a distance measure.

\subsubsection*{Evaluation of Shared Embeddings} 

For those cases where the low-dimensional embedding $f(X) \neq f(X')$, a quantification of the difference between the embeddings is desirable.  A geometric distance measure may be used to determine whether the target embedding $f(X')$ and surrogate embedding $f(X)$ overlap on a global scale.

In this work we utilize the \textbf{normalized symmetric Hausdorff distance}~\cite{taha2015efficient}, which is a geometric measure of the scaled maximum distance possible between points of two datasets, with smaller distances implying greater spatial overlap between two sets of points $D_1$ and $D_2$. The symmetric Hausdorff distance is obtained using $H(D_1, D_2)=max\{\hat{H}(D_1, D_2), \hat{H}(D_2, D_1)\}$, where $\hat{H}(D_1, D_2)= max_{X\in D_1}\{ min_{Y\in D_2}\ \{ \| X, Y \|\} \}$ is the directed Hausdorff distance for each point $X \in D_1$ and $Y \in D_2$, and $\|X,Y\|$ is the Euclidean distance between $X$ and $Y$~\cite{taha2015efficient}. This distance is then normalized by the maximum diagonal of the embedded data so that the maximum distance measurable is 1; a distance of 1 represents the least overlap between two datasets.

\subsubsection*{Feature representations in and extracted from a model}

Feature representations within a model are captured by the model weights. In 2014, Zeiler and Fergus found that features are represented hierarchically within the network: early layers contained high frequency components of an image such as textures and edges, and later layers capture low frequency components such as repeating units and objects~\cite{zeiler2014visualizing}. The hierarchical feature observation strongly influenced the field of explainable AI by establishing the evaluation of network feature representations in the context of how one layer relates to the next~~\cite{li2021visualizing,mostafa2021visualizing,grun2016taxonomy,samek2017explainable, liu2016towards, kahng2017cti, choo2018visual, samek2019towards, nguyen2019understanding, yin2017fine, omeiza2019smooth, qin2018convolutional, zurowietz2020interactive}, and this intuitive understanding of feature representation in a CNN has persisted without major modification~\cite{olah2017feature, NEURIPS2023_76d2f8e3, shaham2024multimodal}. 

However, these approaches do not apply to black-box models, where the weights and architecture are not accessible for manipulation.  In this work, we leverage the feature vector returned by querying the feature extraction backbone.  Although this falls short of a traditional query-based black-box attack, it allows for the search of shared feature representations between models without knowledge of model architecture, weights, and training. Instead of direct inspection of the model weights, the choice of dataset determines which model features are exposed for inspection in the feature vectors returned by the query.


\subsection{Criteria for method that predicts TBA Success}

We propose that a predictive method for the success of black-box TBA should meet standards of directionality and sensitivity~\cite{deng2016data}, in addition to being suitable for the problem of attacking black-box models.  Accordingly, a good \textbf{predictive method} for TBA success will: 
\begin{itemize}
    \item Assume minimal information about the target and surrogate models used for an attack. This requires the method to be agnostic regarding information about training, model weights, gradients, model architectures, etc.
    \item Assume minimal information about the attack.  This includes omitting knowledge about how the attack will be performed (e.g. transfer gradient-based or query-based).
    \item Assume minimal information about the problem domain. The method should only require high-level knowledge of a computer vision task, a natural language processing task, etc.
    \item Predict success and failure equally.  The method should return only true positives and true negatives, satisfying the principle of directionality.
    \item Appropriately differentiate between strong and weak success for a given attack by being sensitive to changes in the model.
\end{itemize}

A method which meets these criteria can be expected to generalize well, and would provide the ability to predict model vulnerabilities to an attack.

\section{Experiments}

\subsection{Method for Identifying Shared Feature Representations}

The procedure identifying shared feature representations is depicted in Figure~\ref{fig:results_siscore} and is summarized as follows:

\begin{enumerate}
  \item Select a test dataset that contains the features a practitioner wishes to query from the model.
  \item Select two or more CNN feature extraction backbones (such as ResNet~\cite{he2016deep}, MobileNet~\cite{howard2019searching}, etc.) for comparison: one as a surrogate backbone and the other as the target backbone.
  \item Extract a set of feature vectors $D_T$ and $D_S$ from each backbone using the test dataset.
  \item Create a low-dimensional manifold embedding from the feature vectors $D_S$ extracted from the reference backbone using a dimensionality reduction algorithm $f(\cdot)$.
  \item Project feature vectors $D_T$ extracted from the target backbone onto the same low-dimensional manifold using the dimensionality reduction algorithm $f(\cdot)$ developed for $D_S$. 
  \item Apply an appropriate manifold analysis.
\end{enumerate}

In Figure~\ref{fig:results_siscore}, a single dataset is used to sample the latent space manifolds of two feature extractors, resulting in two sets of feature vectors, $D_S$ and $D_T$, which encode the same features differently.
The difference in feature encoding is examined by embedding both sets of feature vectors in a single manifold; this manifold is approximated from one of the two sets of feature vectors $X\in D_S$ and $X'\in D_T$ via a dimensionality reduction algorithm such as UMAP~\cite{mcinnes2018umap}.
Metrics are presented which quantify the overlap of the projections.

\subsection{Selection of the test dataset(s)}
\label{ssec:datasets}
Three benchmark datasets were used to probe network performance in different domains, with the implicit assumption that each benchmark has a feature distribution not captured by image instances shared with other datasets~\cite{li2020object, tsipras2020imagenet, cassidy2022analysis}.  For this work, SI-Score~\cite{djolonga2020robustness}, Fashion-MNIST~\cite{DBLP:journals/corr/abs-1708-07747}, and NWPU-RESISC45~\cite{Cheng_2017} were selected.

\textbf{SI-Score}~\cite{djolonga2020robustness} is a synthetic dataset with semantic labels designed to test network robustness to object rotation, size, and location. Additionally, objects in SI-Score have been placed on backgrounds which may or may not be commonly found in nature, therefore implicitly testing the inductive bias of a feature extractor. Because the dataset shares the same classes as the original ImageNet dataset used to train the models, there is a strong assumption that the dataset features are present within the model weights without explicitly testing the models using ImageNet.

The \textbf{Fashion-MNIST dataset}~\cite{DBLP:journals/corr/abs-1708-07747} is a benchmark designed to be as accessible as MNIST~\cite{lecun1998gradient}, but more challenging for object classification tasks. The images were originally 28$\times$28 pixels in size and grayscale, but are resized to $\approx 224\times 224$ pixels to be compatible with the backbone input sizes. Input images are expected to be blurred by the interpolative process, and are therefore dominated by lower frequencies.

The \textbf{NWPU-RESISC45 dataset}~\cite{Cheng_2017} contains large variations in translation, spatial resolution, view point, object pose, illumination, background, and occlusion. Importantly, the image size is fixed to 256$\times$256 pixels, while the image spatial resolutions vary from 30 m to 0.2 m per pixel. This suggests that there is a broad distribution of frequencies present within this dataset that will stress the models' feature representations at multiple scales.  This dataset label space is also completely disjunct from the ImageNet dataset used to train the models, removing any potential intuition regarding feature presence or representation within the model weights.

Details are shown in Technical Appendix Table 2. The number of images was chosen to enable timely processing given available computing resources.

\subsection{Selection of black box models}
\label{ssec:models}
The five feature extractor backbones considered in this study demonstrate a progression of approaches in feature extractor design. ResNetv2~\cite{he2016deep, he2016identity} and MobileNet~\cite{howard2019searching} use residual blocks and inverted residual blocks, respectively. EfficientNet B0 is the result of modifying the MobileNetv2 architecture~\cite{sandler2018mobilenetv2} with the EfficientNet optimization process ~\cite{tan2019efficientnet}. The primary differences between MobileNetv2 and MobileNetv3 are the removal of computationally expensive layers, substitution of the \textit{hard sigmoid} instead of \textit{ReLU}, and implementation of squeeze and excitation in MobileNetv3~\cite{howard2019searching}. Inception-v3~\cite{szegedy2016rethinking, incep_tfhub} and Inception-ResNet-v2~\cite{szegedy2017inception, incepRes_tfhub} differ in the inclusion of batch normalization, residual connections, and filter concatenation. 

The models were implemented through the TensorFlow Hub model library: each model was previously trained using the ILSVRC 2012 ``ImageNet'' dataset~\cite{ILSVRC15} and was not fine-tuned.  Additional implementation details are presented in Technical Appendix Table 1.

\subsection{Transfer-based Attack Analysis}

A classic Fast Gradient Sign Method (FGSM) attack~\cite{goodfellow2014explaining, muncsan2021transferability} with the form $x^* = x + \epsilon \cdot \textrm{sign} \left( \nabla_xJ\left(\theta, x, y \right) \right)$ is used to generate attacks for each model, where $x$ is the original image, $x^*$ is the adversarial image, $y$ is the original label, $\epsilon$ is the attack strength, $J(\cdot)$ is the loss, and $\theta$ signifies the model parameters.  Each feature extraction backbone initialized with ImageNet weights was connected to a classification head consisting of a drop out layer, followed by a fully connected layer; the backbone weights were frozen, and only the classification head was trained.  Each model was then trained until the training and validation losses converged.  The models each achieved an average F1 score of $0.9 \pm 0.1$ on a held-out test set except for IncepNetv3 \cite{szegedy2016rethinking, incep_tfhub}, which had a maximum F1 score of $\approx 0.6$.  

Each model was then attacked using the gradients generated from the FGSM~\cite{goodfellow2014explaining} attack with increasing attack strength $\epsilon$ and evaluated using the F1 score and the average accuracy (AA).  The AA for each attack at $\epsilon$ threshold $\epsilon = \ell_\infty=8/255=0.03$~\cite{li2024towards} is also reported.  A lower AA value indicates a more successful attack.

\section{Results}
\label{ssec:usefri_results}

In this section, the method of Figure \ref{fig:results_siscore} is demonstrated using MobileNet v3 as the target model, and ResNet50 v2 as the surrogate model. Additional results for other surrogate-target model pairs are presented in the Technical Appendix.  The results of all surrogate-target model pairs are summarized in the main result shown in Figure \ref{fig:attack_vs_H}, which demonstrates the correlation between embedding similarity and attack success.

\subsection{Shared feature vector embeddings generated by cross projection }
\label{sssec:umap_proj}
\begin{figure}[!ht]
    \centering
    \includegraphics[width=0.48\textwidth]{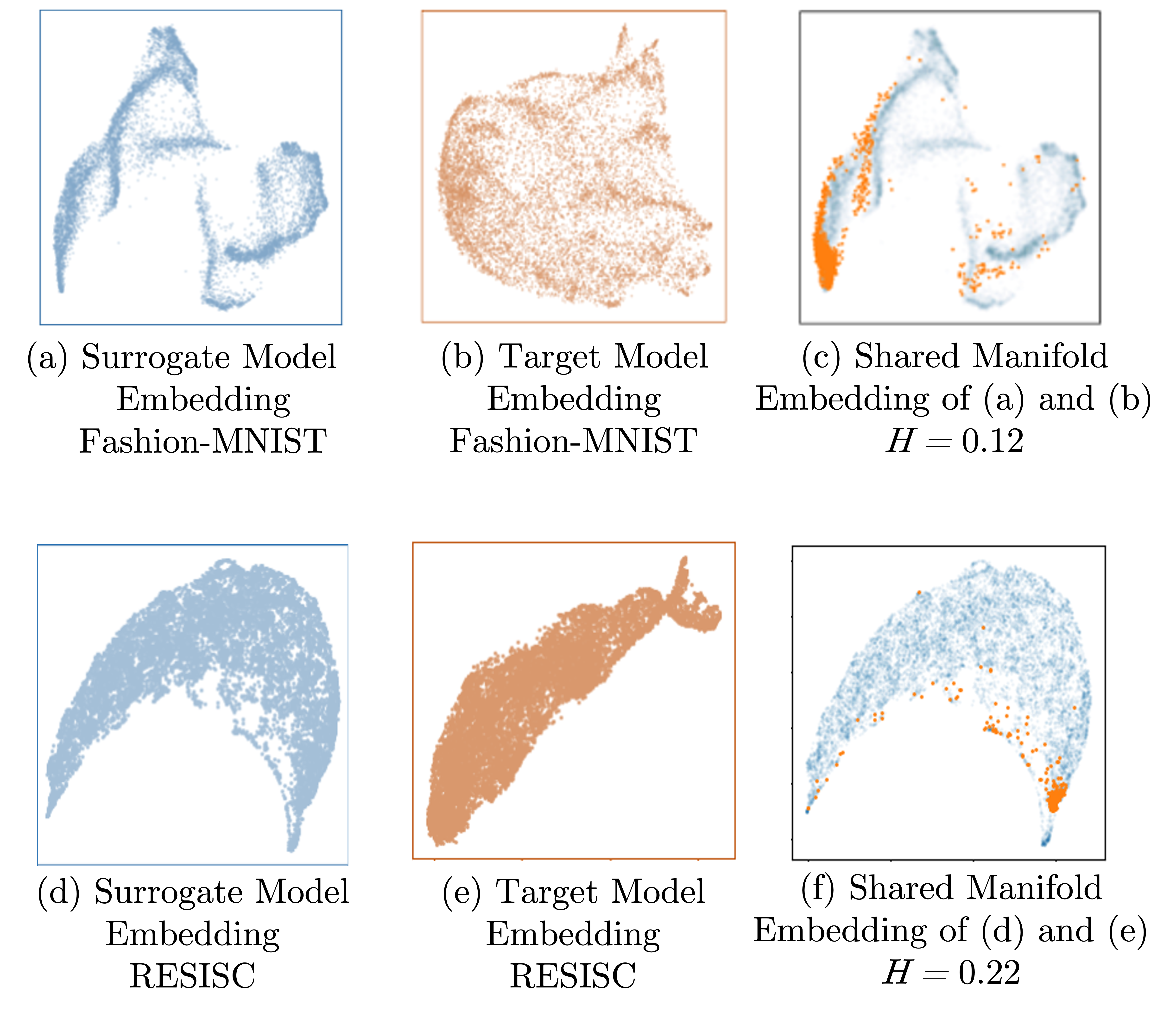}
    \caption{Analysis of manifold embeddings for Fashion-MNIST and NSWC-RESISC45 ``RESISC''.  (a), (d) The surrogate model embedding generated using feature vectors $X$ from ResNetv2 and embedded by $f(X)_{UMAP}$.  (b), (e) The target model embedding, generated using feature vectors $X'$ from MobileNetv3 and embedded by UMAP $g(X')_{UMAP}$. (c), (f) cross manifold embedding generated by transformation $f(X')$ and plotting with $f(X)$.  The Hausdorff distance between the two embeddings is shown, where a larger value of $H$ implies that the embeddings overlap less. }
    \label{fig:results_fashion}
\end{figure}

The cross manifold embeddings of the SI-Score dataset are shown in Figure \ref{fig:results_siscore}.  The Fashion-MNIST and RESISC feature vectors from ResNetv2 50 and MobileNetv3 are shown in Figure \ref{fig:results_fashion}.  
When data from the target model clusters tightly on the surrogate manifold projection, only a few neighbors were found in the original surrogate high-dimensional space for the entire set of target feature vectors. This implies that, although every data point was projected successfully (i.e., no NaN values), the high-dimensional feature spaces generated by the target and surrogate models were too disparate to survive the cross projection process. The result is a shared embedding with a Hausdorff distance that approaches $H=1$ as fewer neighbors are found. For projections that share many neighbors between the target and surrogate embeddings, the Hausdorff distance approaches zero.  This is visualized in Figure \ref{fig:results_fashion}(f) for RESISC data, with $H=0.22$, Figure \ref{fig:results_fashion}(c) for Fashion-MNIST data with $H=0.12$, and Figure \ref{fig:results_siscore} for SI-Score with $H=0.09$.

Additional embeddings are presented in the Technical Appendix for the remaining model and data combinations.

\subsection{Application to Transfer-based attacks}
\label{ssec:blk_box_attacks}

\begin{figure}[!ht]
    \centering
     \includegraphics[width=0.45\textwidth]{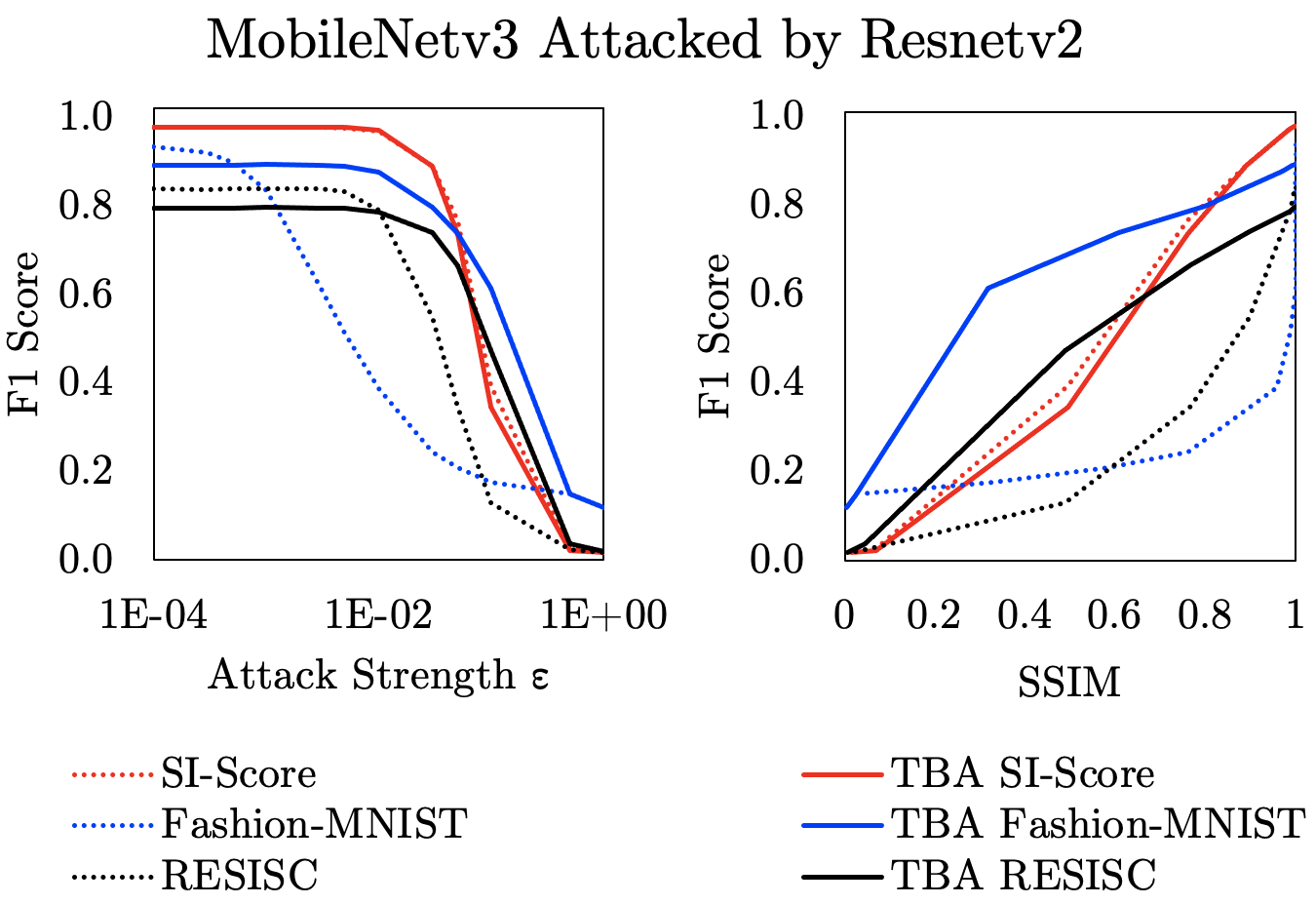}
    \caption{Comparison of FGSM attacks generated by the ResNetv2 surrogate model against itself (solid line) and against the target MobileNet v2 model (dashed line) for different source datasets SI-Score (red), Fashion-MNIST (blue), and RESISC (green).  (LEFT) Model performance for attack strength $\epsilon$. (RIGHT) Model performance for perturbation of the input images as evaluated by the average SSIM of the attacked images. Error bars are not shown, as the standard deviation is on the order of 1E-3 to 1E-6.}
    \label{fig:attacks}
\end{figure}

The value of understanding latent space similarity is apparent in the context of transfer-based attacks (TBA).  In Figure \ref{fig:attacks}, results for FGSM attacks created by the ResNetv2 surrogate model for the target MobileNetv3 model are shown.  The dashed lines represent how the attack performed against the surrogate model for which it was created, and the solid line represents how well the attack transferred to the target model for various strengths.  The difference between the dashed line and solid line of a single color represents the change in attack success when applied to the source surrogate model (dashed line) and the target model (solid line).  

The SI-Score dataset (red) required the strongest FGSM attack with an attack strength of $\epsilon = 1E-01$ to achieve an initial drop in accuracy, but the attack performed equally well against the target and surrogate models.  The TBA generated with ResNetv2 and SI-Score has a success of $AA(\epsilon = 0.03) = 0.89$ on our MobileNetv3 model, representing a 10\% decrease in performance for MobileNetv3 on this dataset; the corresponding Hausdorff distance as shown in Figure \ref{fig:results_siscore} is $H=0.09$. These results can be compared with the attack using the Fashion-MNIST dataset (blue), which required the strength of the attack $\epsilon$ to be increased by an order of magnitude to achieve the same success on the target model.  For the Fashion-MNIST attack against the target model, $AA(\epsilon = 0.03) = 0.79$ representing an 11\% decrease in model performance on the given classification task; the Hausdorff distance for this data as shown in Figure \ref{fig:results_fashion}(c) is $H=0.12$.  The attack generated using RESISC data (black line) also requires the attack strength to be increased by an order of magnitude to achieve similar success between the surrogate and target models: for the target model, $AA(\epsilon = 0.03) = 0.74$ represents a 7\% decrease in model performance.  

The same attacks are also reported in terms of image perturbation in Figure \ref{fig:attacks}, where the perturbation is quantified using the average SSIM score~\cite{bakurov2022structural}.  Curves which approach the upper left corner represent models robust to the attack.  The SI-Score dataset again shows the greatest similarity between surrogate and attack model performance, while the remaining two datasets diverge in their transferability. 

\begin{figure}[!ht]
    \centering
    \includegraphics[width=0.48\textwidth]{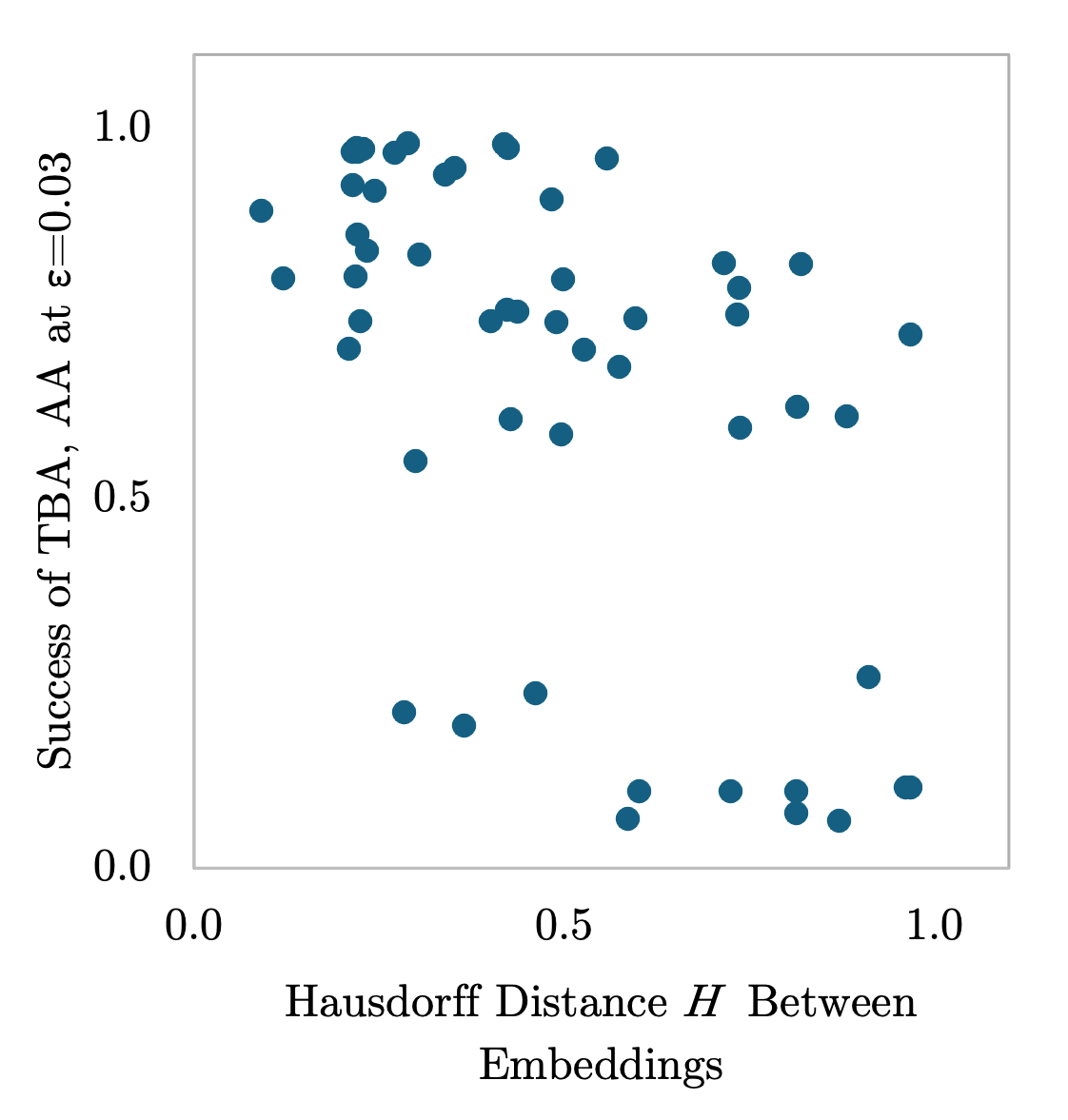}
    \caption{The success of TBA as quantified by the average accuracy \textit{AA} at attack strength $\epsilon=0.03$ plotted against the normalized symmetric Hausdorff distance between the target and surrogate data embeddings on a shared low-dimensional manifold.  An inverse correlation is observed, where successful attacks (high AA) correlate with small Hausdorff distances. The correlation coefficient $\rho_{H,AA}$is $-0.56$, indicating a moderate negative correlation between the distance $H$ beteween embeddings and the average accuracy $AA$ of the model.  Analyzing this data using PCA results in eigenvalues of $\left[ 7.13, 0.00\right]$, indicating that the data variance can be explained by a single direction.
    }
    \label{fig:attack_vs_H}
\end{figure}

This analysis was repeated for different surrogate-target-dataset combinations.  The AA of the TBA at $\epsilon=0.03$ and the Hausdorff distance between surrogate and target embeddings are shown in Figure \ref{fig:attack_vs_H}.  The data for this plot, along with details of datasets and the surrogate-target model combinations, are included in the Technical Appendix. 

A negative correlation is observed, with small Hausdorff distances between embeddings correlating with improved success rate, and Hausdorff distances approaching the maximum correlated with decreased success.  This is quantified by the correlation coefficient $\rho_{H, AA} = E\left[ \left(H - \mu_H \right) \left(AA - \mu_{AA} \right) \right]/\left(\sigma_H \sigma_{AA} \right) = -0.57$ for the set of Hausdorff distances $H$ and the corresponding average accuracy $AA$.  Applying Principal Component Analysis (PCA) \cite{greenacre2022principal} to the data in Figure \ref{fig:attack_vs_H} yields eigenvalues $\left[ 7.13, 0.00\right]$. This indicates that the data variance can be explained by a single component of the PCA projection.
\section{Discussion}

These results support the manifold attack theory, where the similarity--or non-similarity--of feature representations motivates vulnerability to transfer-based attacks (TBA)~\cite{chen2024theory}.  The correlation observed in Figure \ref{fig:attack_vs_H} is moderate, but present.  The following discussion identifies key analysis of the results, and directions for future work.

\subsection{Dimensionality reduction algorithm for generating shared embeddings}
\label{ssec:dim_red_algo}
The best choice of dimensionality reduction algorithm for this method is an open question.  The projection of data from one metric space onto another space requires some mathematical assumptions not discussed in detail here, namely the requirement that the spaces must be isometric with a bijection that preserves distances~\cite{chazal2021introduction}. The dimensionality reduction algorithm for this method should meet these criteria.

Here, the criteria are assumed to be satisfied in the high-dimensional space of the feature vectors extracted from the CNN backbones, and are supported by empirical results (preservation of some structure during cross embeddings) following the convention of other authors in the field~\cite{barannikov2021manifold, you2022comparing, liu2019latent}. However, a more rigorous analysis would explicitly determine the Gromov-Hausdorff distance~\cite{edwards1975structure, gromov1981groups} between the various high-dimensional spaces, as well as between UMAP projections, before attempting additional metrics such as the Hausdorff distance. 

During implementation, this work resolved differences in dimensionality between the feature spaces of the target and surrogate models by zero-padding the length of the feature vector to the maximum dimensionality of 2048 dimensions. This is justifiable because 1) it preserves all information of the larger-dimensioned feature space, 2) preserves the spatial relationship between values in each feature vector as encoded by the output of various CNN kernels within the feature extraction backbone, and 3) makes use of the same assumption of shared ``lower-dimensional" winning lottery ticket as the rest of this work: if both the target model and surrogate model organize information similarly, then the smaller-dimensioned feature vector should contain the same information organized in the same way as the larger-dimensioned feature vector.  The zero-padding can be considered to restore higher-dimensions which are unused in the smaller feature vector representation.

The alternative to zero padding--truncating feature vectors to a minimum shared dimensionality either by cropping or by enforcing significance in all values of the vector and removing small values--overlooks permutations that preserve spatial relationships, such as cyclical shifts where the shared subspace is in the last 1280 values rather than the first. Compared to truncation, zero-padding may increase distance during the $k$NN computation, but it will not falsely minimize it. 

Importantly, if the assumption of similar feature organization is \textit{incorrect}, feature vectors will fail to be neighbors in the kNN embedding of the UMAP algorithm, and the Hausdorff distance will not accurately predict shared feature representations or the success of a transfer-based attack. Our proposed method does not artificially create similar feature representations where none exist, which helps to reduce the likelihood of false positives and ensures that the method adheres to the criteria established earlier in this work. 

Future work should identify the subspace that minimizes differences between target and surrogate representations to avoid false negatives. A stronger correlation between the cross manifold embeddings and the attack success could well be observed by performing such a subspace alignment between latent spaces with differing dimensions. This approach would maintain the method's validity without introducing forced similarities. Additionally, while UMAP hyperparameters do influence the structure of dimension-reduced embeddings, manifold features remain consistent. Further details on UMAP stability are provided in the Technical Appendix. 

\subsection{Choice of similarity evaluation approach for embeddings} 

The normalized symmetric Hausdorff distance~\cite{taha2015efficient} quantifies whether the two data projections $f(X)$ and $f(X')$ are similar on a global scale.  This metric fails to consider whether the projections are similar on a local scale.  One such measure of local topology is the bottleneck distance~\cite{cohen2005stability}; sample calculations of bottleneck distances for MobileNetv3 and ResNetv2 projections are included in the Technical Appendix.  Although the bottleneck distance has the desirable property of comparing the stability of data clustering at all scales between the two projections~\cite{cohen2005stability}, in practice we found the computational burden to outweigh the potential benefits of the method.  The bottleneck distances calculated were also not strongly predictive of the success of transfer-based attacks, and had no immediate benefit over the reported Hausdorff distances.

\subsection{Correlation of TBA Success and Shared Feature Representations}

In Figure \ref{fig:attack_vs_H}, a negative correlation between the Hausdorff distance of the shared embeddings and the success of a transfer-based attack is observed.  This trend is not linear, and attempts to fit a linear model resulted in an $R^2 = 0.32$ due to the data variance. This variance can be partially attributed to the assumptions imposed on the embeddings, as discussed above in \S \ref{ssec:dim_red_algo}.

The variance may also be attributed to the observed failure of the FGSM attack implemented here to optimally transfer between models~\cite{li2024towards}. In Figure \ref{fig:attack_vs_H}, the data is linearly separable at $AA=0.4$ indicating models which are generally robust to attack ($AA > 0.4$) and models which are vulnerable to attack ($AA < 0.4$). The use of an improved TBA method is expected to result in an increased attack success rate and a decreased average accuracy $AA$ value.  

Given a reliable shared embedding method, this method proposed in this work is expected to be sensitive to when a TBA method is \textbf{not} motivated by a shared lower-dimensional manifold of feature representations in the backbone. An example of this sensitivity is explained by shared feature representations in the CNN backbone may be found by considering the classification prediction head of the model. The prediction head was omitted from the similarity analysis between the surrogate and target models, as it does not return information regarding feature representation in the backbone, and this is the topic of our work. However, the classification head's impact is still included in the accuracy scores for each attack, and therefore effects the results reported in Figure \ref{fig:attack_vs_H}. Therefore, this method could return a false negative when similar feature representations in the target and surrogate models are obscured by a difference in how those features are used in predictions. Future work should implement methods~\cite{wei2016networkmorphism,rashno2022uncertainty} for quantifying the uncertainty a prediction head introduces to a TBA-success model predicated on feature representation within the model backbone.
\section{Conclusion}

A primary contribution of this work is the identification and quantification of similarity in feature representations within black-box models through cross manifold embedding, quantified by the normalized symmetric Hausdorff distance. Previous methods required direct inspection of model parameters~\cite{athiwaratkun2015feature} and interpretation of weight values~\cite{bayar2017augmented}. This work avoids such evaluations by placing constraints on the models' latent spaces and the dimensionality reduction techniques used to compare them. Future work should explicitly verify that these constraints are satisfied using methods such as the Gromov-Hausdorff distance~\cite{edwards1975structure, gromov1981groups} or methods yet to be developed.

However, even with unverified assumptions, decreased similarity in feature representation is negatively correlated with increased success of the transfer-based attack by correlation coefficient $\rho=-0.56$ for a FGSM attack with fixed attack strength $\epsilon = \ell_{\infty}$.  Because the methodology introduced in this work does not require knowledge of the transfer-based attack, the surrogate model or test model, it becomes possible to predict whether a target-based attack will succeed on a black-box model. The predictive methodology presented here can be further improved by considering the uncertainty introduced by the model prediction heads, leveraging additional attack success evaluation approaches and feature representation similarity, and maximizing the shared semantic information between the latent spaces of target and surrogate models.

We encourage the TBA community to develop and improve methodologies that predict the success of transfer-based attacks against black-box models and identify vulnerable feature representations, as outlined in this work. Such methodologies would enable the identification of vulnerable models before deployment, increasing the trustworthiness and reliability of systems that leverage machine learning algorithms.

\section*{Acknowledgements}
The authors gratefully acknowledge contributions to experimental design by William Boler.  This work was funded by Naval Surface Warfare center for Broad Agency Announcement for FY18 SOLICITATION NO.: DOTC INIT0181, and from ONR CDEW: Contract\textbackslash{}Grant\textbackslash{}MIPR \# – N00014-20-1-2674, the US Department of Defense (Contract W52P1J2093009), NSWC-CRANE N00164-24-1-1001.

{
    \small
    \bibliographystyle{ieeenat_fullname}
    \bibliography{main}
}

\clearpage
\setcounter{page}{1}
\setcounter{section}{0}

\maketitlesupplementary
\onecolumn
\section{Technical Appendix and Supporting Results}
Note: Additional references may be found at the end of this section.

\subsection{Data and Model Details}

\begin{table*}[!ht]
 \caption{Feature extractor networks selected for analysis}
 \label{tab:networks}
 \centering
 \begin{tabular}{p{0.15\linewidth}p{0.15\linewidth}p{0.08\linewidth}p{0.12\linewidth}p{0.15\linewidth}}
 \toprule
 \bf Model Name & \bf Num. Wts. & \bf Feat. Vec. Size & \bf Year & \bf Paper Citation \\
 \midrule
 ResNetv2 50 (w/ full preactivation) & 23,564,800 & 2048 & 2015-16 &~\cite{he2016deep,he2016identity,resnet_tfhub} \\
 MobileNet V3 large     & 4,226,432  & 1280 & 2019 &~\cite{howard2019searching, mobile_tfhub} \\
 EfficientNet B0      & 4,049,564 & 1280 & 2019 &~\cite{tan2019efficientnet, effNet_tfhub} \\
 Inception-v3            & 21,802,784  & 2048  & 2016 &~\cite{szegedy2016rethinking, incep_tfhub} \\
  Inception-ResNet-v2         & 54,336,736  & 1536 & 2017 &~\cite{szegedy2017inception, incepRes_tfhub}   \\
 \bottomrule
 \end{tabular}
\end{table*}

\begin{table*}[!h]
 \caption{Datasets selected for analysis}
 \label{tab:datasets}
 \centering
 \begin{tabular}{llll}
  \toprule
  Dataset Name  & Number of Classes & Number of Images & Paper Citation \\
  \midrule
  SI-Score    & 1000       & 39,540 &~\cite{djolonga2020robustness} \\
  Fashion-MNIST  & 10        & 10k (test set) &~\cite{DBLP:journals/corr/abs-1708-07747} \\
   NWPU-RESISC45 & 45    & 10k sampled from 31,500 &~\cite{Cheng_2017}  \\
  \bottomrule
 \end{tabular}
\end{table*}

\subsection{Implementation Details}
\label{ssec:experiments}
A set of feature vectors $X_S$ were extracted from each backbone using the source dataset $D_S$. After extraction, the sets of feature vectors for each backbone were fixed, and used repeatedly throughout the various experiments.

Dimensionality reduction to 2D was performed on each set of feature vectors using UMAP with hyperparameters \texttt{num\_neighbors} $=[20, 50, 500]$ and \texttt{distance} $=[0.25, 0.1, 0.1]$ for SI-Score, RESISC45, and Fashion-MNIST respectively. Other hyperparameters were left to the default values. Results are shown in Figure \ref{fig:results_fashion} of the paper. Once the visualized data was observed to have some structure, no further attempts were made to fine-tune hyperparameters based on class labeling.

For the results visualized in Figure \ref{fig:results_fashion} of the paper, the ResNetv2 UMAP embeddings were chosen to serve as the surrogate model. Additional results with different surrogate models are shown below.  The feature vectors from target backbones were then transformed using the ResNetv2 UMAP embedding of that dataset using the \texttt{transform} function, a process referred to as \textit{cross projection}. No projected data points were assigned a NaN value, indicating that every cross-projected data point found a neighbor in the original kNN initialization for the surrogate UMAP embedding~\cite{umap}.

All data generation and topological data analysis used a single NVIDIA RTX A6000 with 48 GB RAM. The TensorFlow Hub and API~\cite{tf} were used for the feature extraction process. 

\subsection{Validity of UMAP as Dimensionality Reduction Algorithm}

\subsubsection{Are the UMAP embeddings repeatable?}

Because the UMAP algorithm implements a stochastic optimization step, it was worth considering whether the UMAP embeddings are repeatable.  Small variations in distribution can be quantified using Procrustes Analysis~\cite{procrustes}. 

The disparity results from ten ResNetv2 UMAP embeddings are shown in Table \ref{tab:procrustes}.   To determine the disparity between repeated UMAP trials, the second through tenth projections from the UMAP embedding process were compared with the first projection using Procrustes analysis.  No uncertainty is reported for the values: each repeated Procrustes analysis returned the same disparity value $M$ nine times for a given dataset.  Procrustes analysis repeated for the same projection (self-comparison) correctly returned a value of zero.

\begin{table}[!h]
  \caption{Disparity results for repeated ResNetv2 50 UMAP embeddings}
  \label{tab:procrustes}
  \centering
  \begin{tabular}{lccc}
    \toprule
     &   FashionMNIST&  RESISC45  & SI-Score  \\
    \midrule
    \textbf{M} & 3.520E-16& 4.342E-16  & 1.175E-16   \\
    \bottomrule
  \end{tabular}
\end{table}

\subsubsection{What happens when changing the number of nearest neighbors in the UMAP algorithm?}

\begin{figure}[!h]
    \centering
    \includegraphics[width=0.5\linewidth]{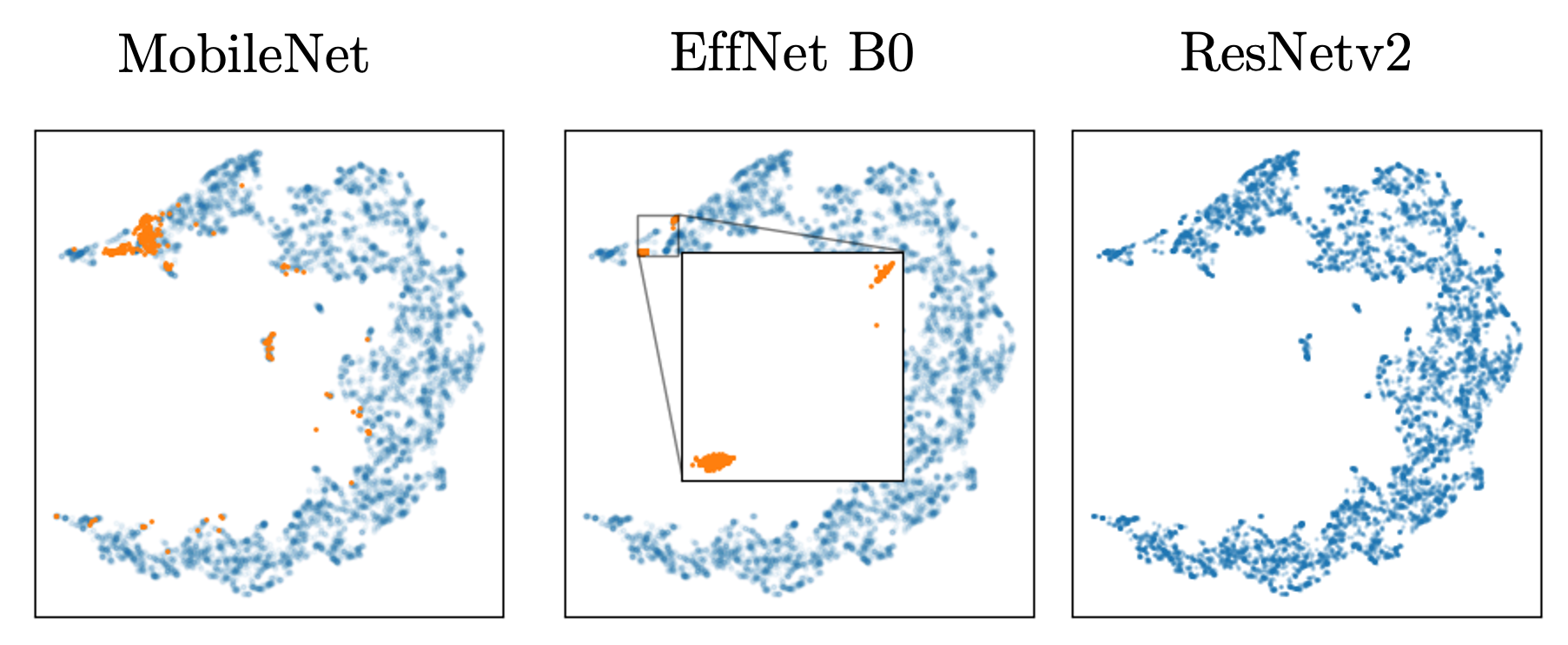}
    \caption{Repeating the UMAP embedding process with new hyperparameters: k=5 and distance=1$\times10^{-10}$ for the RESISC dataset and ResNetv2 surrogate embedding (shown at far right).  The target data continues to have neighbors on the ResNet embedding, although the total area covered by the target dataset is diminished.}
    \label{fig:sup_knn}
\end{figure}

Reducing the number of neighbors in the UMAP algorithm was expected to have an impact on the results, but when tested, we observed the change to be minimal.  This is shown in Figure \ref{fig:sup_knn}, where the ResNetv2 embedding of RESISC dataset was used for targets MobileNetv3 and EffNet.  The hyperparameters were changed from $k=50$ to $k=5$, the distance was changed from $0.1$ to $1\times10^{-10}$.

When compared to Figure \ref{fig:sup_cross_embed}, the MobileNetv3 and EffNet embeddings on ResNetv2 have similar structure.  However, we consider the requirement of any hyperparameter fitting to be a weakness of this work.

\subsubsection{Does this method work for the trivial case?}
\label{ssec:umap_same_model}

In order to test the trivial case where assumed shared latent space representation and similar latent feature vectors are well justified, this section presents a study using Fashion-MNIST~\cite{fashionmnist} and VGG-19~\cite{vgg} models.

\begin{figure*}[!ht]
    \centering
    \includegraphics[width=0.8\linewidth]{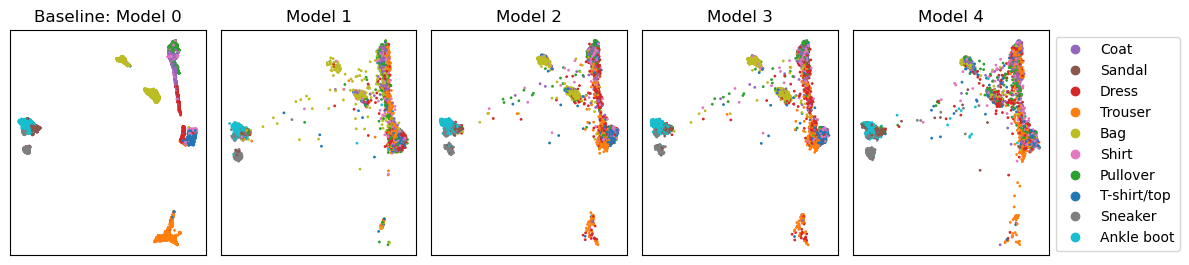}
    \caption{Results from training VGG-19 from scratch five times with Fashion-MNIST, using feature vectors from the zeroth model to create a UMAP transformer, then transforming feature vectors from Models 1-4.  The structure is remarkably preserved, suggesting that the models have learned to represent information similarly and this is captured by the dimensionality reduction.}
    \label{fig:vgg_transforms}
\end{figure*}

Five VGG-19~\cite{vgg} models were randomly initialized and trained from scratch using the Fashion-MNIST~\cite{fashionmnist} train partition of 60k images.  The models were trained using early stopping, with a maximum number of 50 epochs and a learning rate of 0.001.  The models were then evaluated using the Fashion-MNIST test partition, and achieved an average F1 score of $0.88 \pm 0.05$. The feature vectors were extracted from each of the five models, and the feature vectors from the zeroth model were used to train a UMAP data transformer.  UMAP hyperparameters used 10 nearest neighbors, a minimum distance of 0.1, and the cosine distance. The feature vectors from the remaining four models were then transformed using the UMAP trained on the zeroth model feature vectors. The results are shown in Figure \ref{fig:vgg_transforms}.

These results show that although there are some noise differences between  different models, the overall global structure was preserved when using the zeroth-model UMAP to transform other models with similarly structured latent spaces.  This suggests that although the models were trained from scratch each time, the latent spaces organized the features learned from the Fashion-MNIST training set similarly each time.  \textit{\textbf{Importantly, the UMAP algorithm did not substantively create or destroy similarities between the feature representations.}}

\subsection{Original embeddings from different models}

\begin{figure}[!h]
    \centering
    \includegraphics[width=\linewidth]{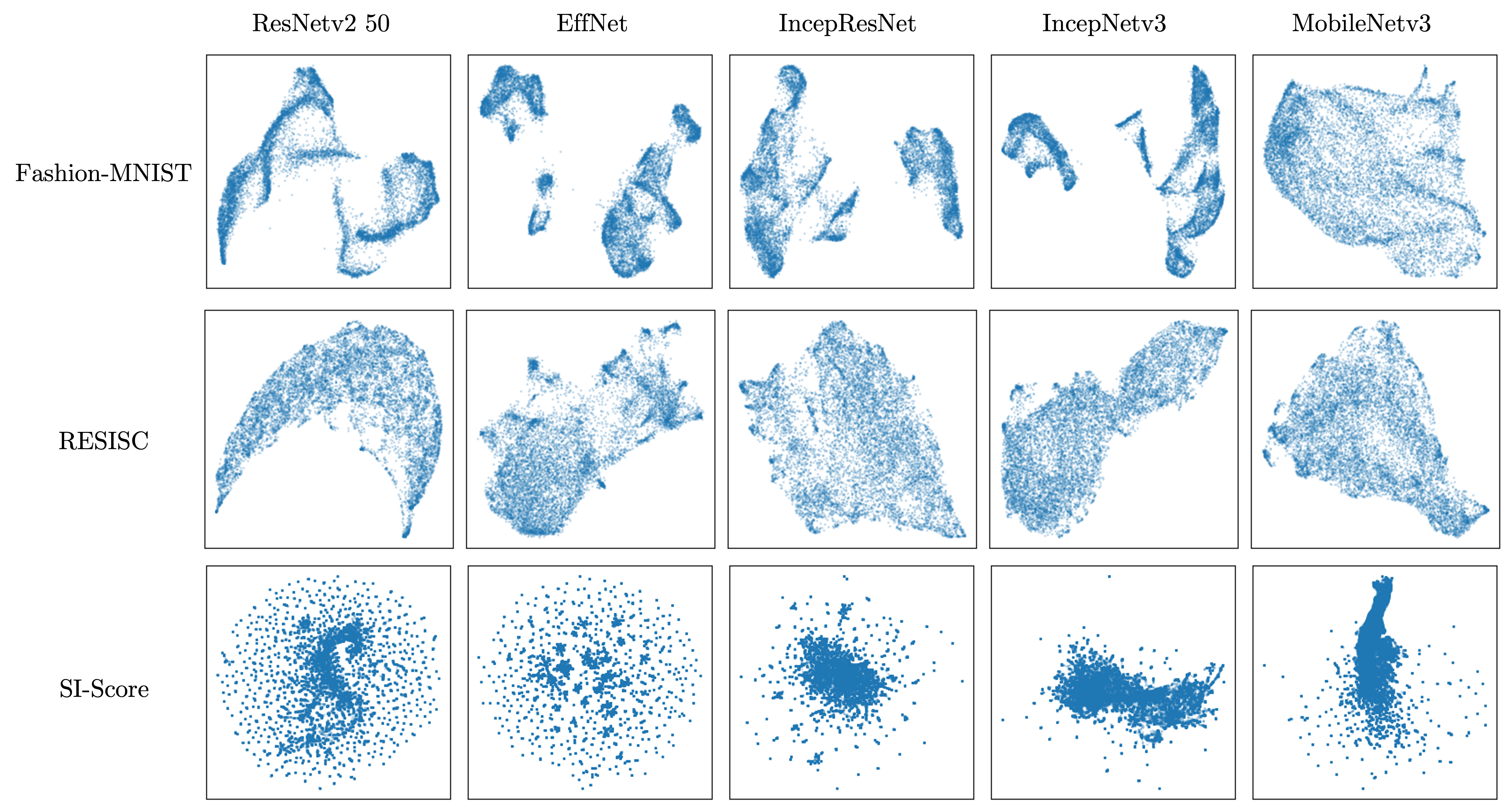}
    \caption{Comparison of native UMAP embeddings of the same dataset from different models using the same hyperparameters for each embedding.  TOP Row: Fashion-MNIST, MIDDLE Row: RESISC, BOTTOM Row: SI-Score.  The same dataset presents differently depending on the model used to generate the embedding.}
    \label{fig:og_embeddings}
\end{figure}

The feature vectors extracted from each model for each dataset are shown in Figure \ref{fig:og_embeddings}.  Each row represents a dataset, each column represents a model.

Each one of these embeddings can be used as the surrogate manifold by implementing the dimensionality reduction function generated for that manifold on additional datasets.  The results of using manifolds from Figure \ref{fig:og_embeddings} as surrogates are presented in the following sections. Hausdorff distances are presented for the distance between each target and surrogate embedding.

\subsection{ResNetv2 as Surrogate Model}
\begin{figure}[!h]
    \centering
    \includegraphics[width=0.7\linewidth]{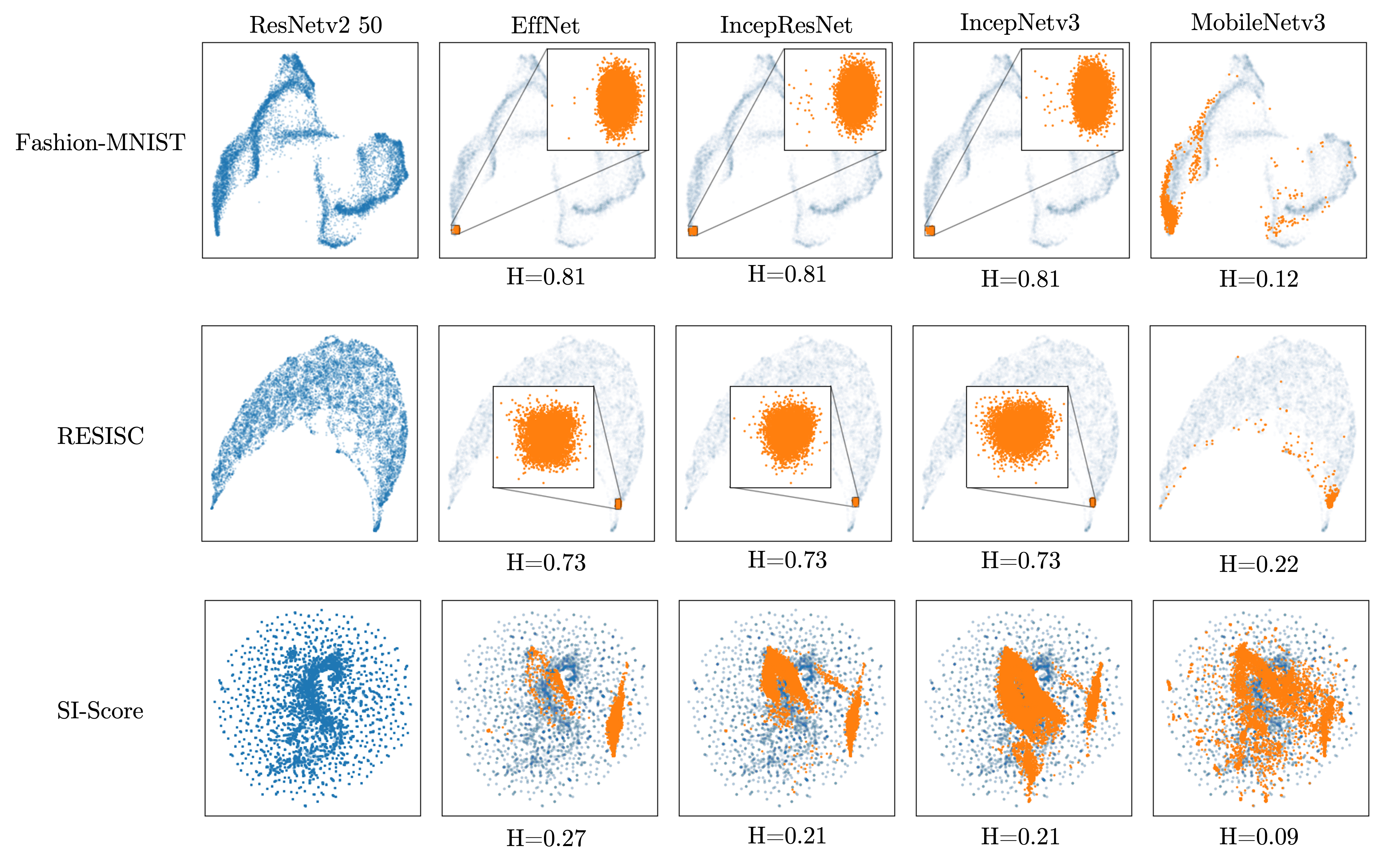}
    \caption{Comparison of projecting dataset $x \in$ [Fashion-MNIST, RESISC, SI-Score] (orange) from a target model onto a surrogate model manifold generated from \textbf{ResNetv2} feature vectors (blue). The projected data from a target model was not consistently similar to the surrogate model embedding, sometimes overlapping well and sometimes collapsing to a small cluster.}
    \label{fig:sup_cross_embed}
\end{figure}

In Figure \ref{fig:sup_cross_embed}, ResNetv2 50 was used as the surrogate model for each dataset and model.  The transfer-based attack results for these surrogate-target model combinations are included in Table \ref{tab:raw_results}. By visual inspection, the MobileNetv3 embeddings share the most structure with ResNetv2, and the SI-Score dataset was most consistently similar across all models.

When comparing differences between Figure \ref{fig:sup_cross_embed} and Figure \ref{fig:sup_mob_resisc}, the algorithm implements eigenvector sets which may be different from a previous embedding with the target and surrogate model roles reversed while still capturing a shared feature structure~\cite{umap}. This can be interpreted as arising from the same features present in both cases, but represented differently depending on the eigenbasis used to create the UMAP embedding.  The interpretability of the features was limited by the basis in which they were expressed.

\subsection{MobileNetv3 as Surrogate Model}
\begin{figure}[!h]
    \centering
    \includegraphics[width=0.7\linewidth]{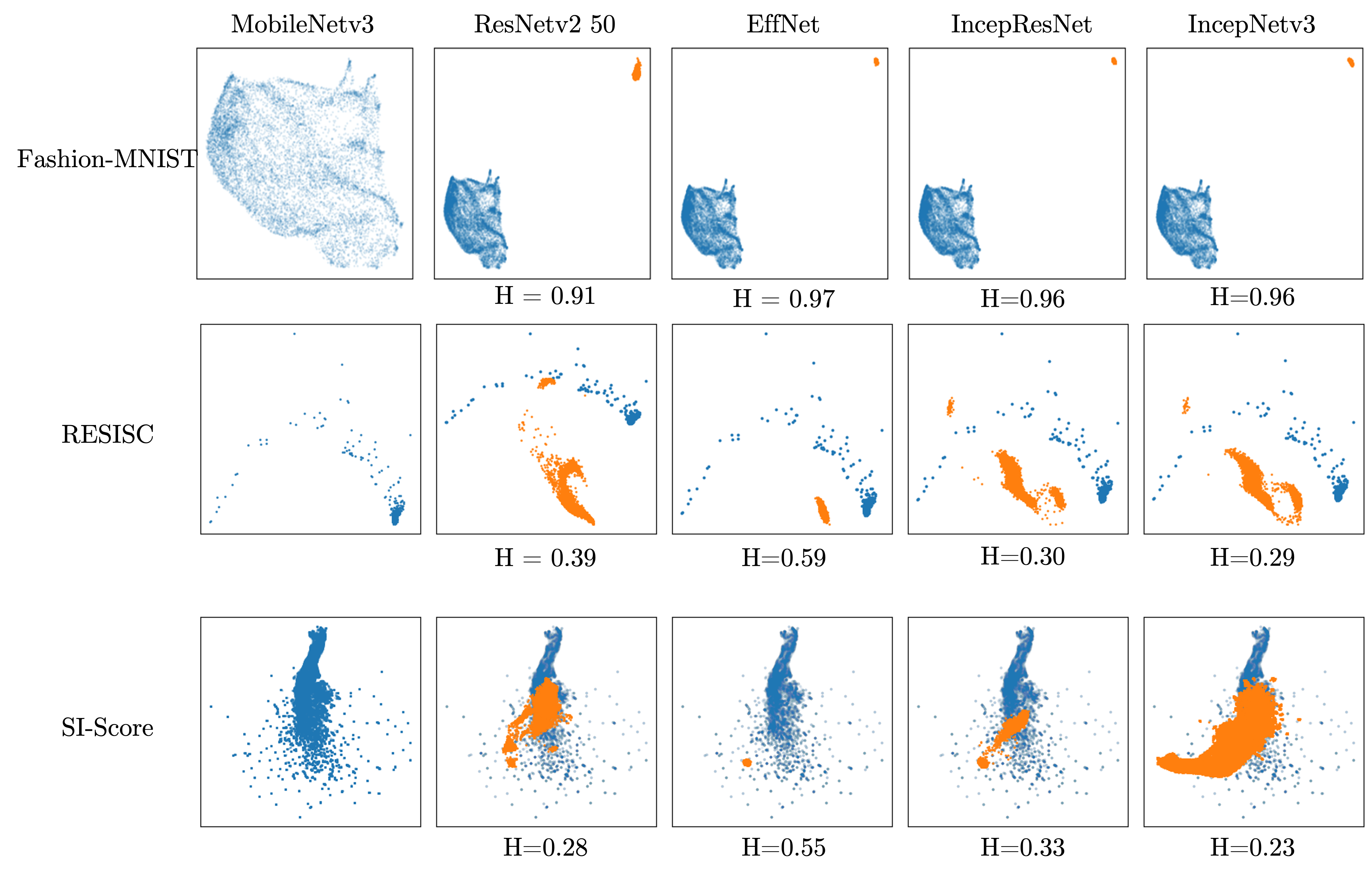}
    \caption{Comparison of projecting dataset $x \in$ [Fashion-MNIST, RESISC, SI-Score] (orange) from a target model onto a surrogate model manifold generated from \textbf{Mobilenetv3} (blue). Hausdorff distances for each shared embedding are shown. The structure of the projected embeddings differs from those shown in Figure \ref{fig:sup_cross_embed}.}
    \label{fig:sup_mob_resisc}
\end{figure}

In Figure \ref{fig:sup_mob_resisc}, MobileNetv3 Large was used as the surrogate model for each dataset and model. When compared to Figure \ref{fig:sup_cross_embed}, ResNetv2 and MobileNetv3 are no longer the most similar embeddings.  The Hausdorff distance continues to be overly sensitive to outliers and not group structure, which motivates the introduction of topological metrics presented below.  The transfer-based attack results for these surrogate-target model combinations are included in Table \ref{tab:raw_results}.

\clearpage
\subsection{Additional Attack Data}

\begin{figure}[!h]
    \centering
    \includegraphics[width=0.5\textwidth]{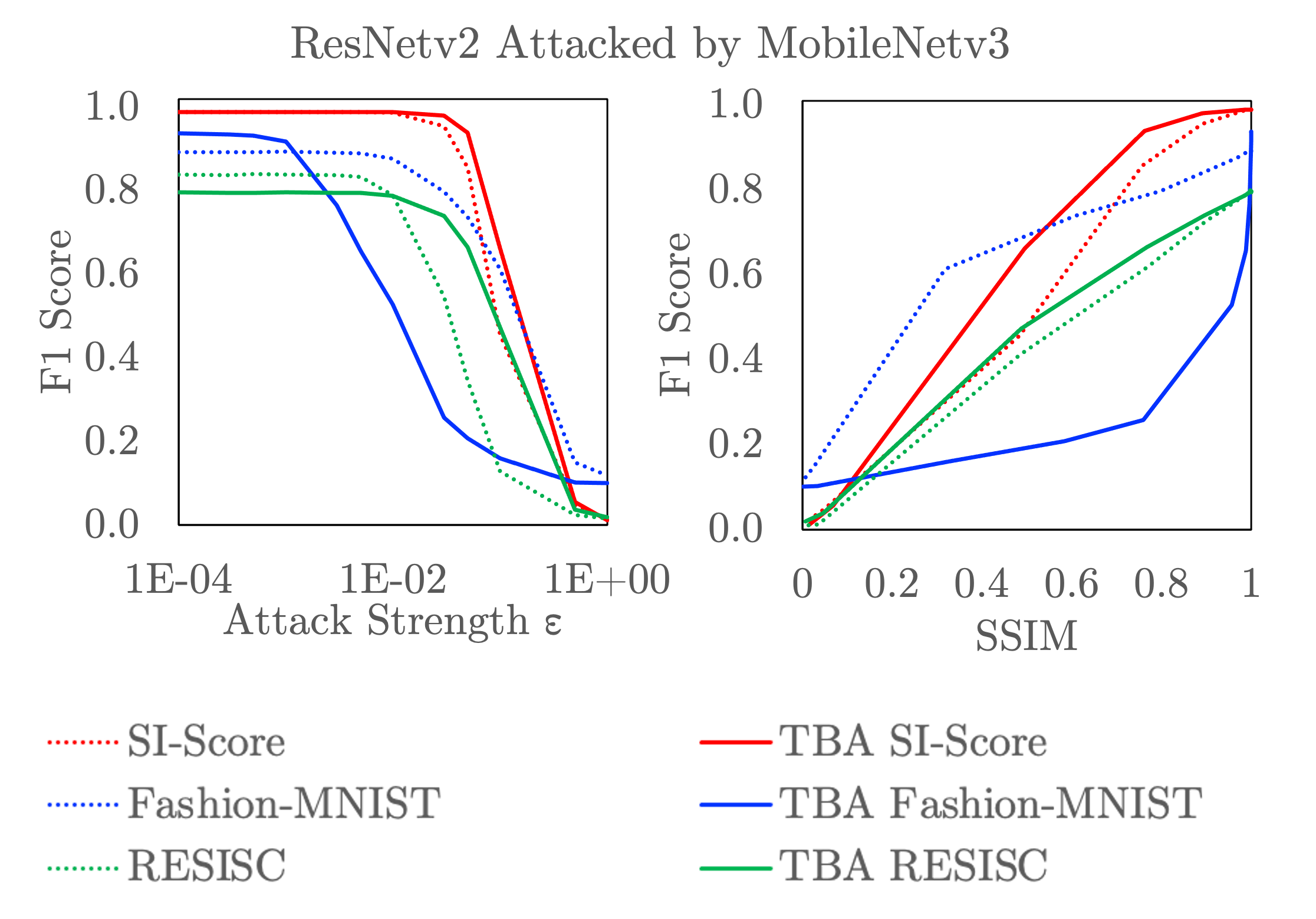}
    \caption{Comparison of FGSM attacks generated by MobileNetv3 as the surrogate model against itself (solid line) and against  ResNetv2 as the target model (dashed line) for different source datasets SI-Score (red), Fashion-MNIST (blue), and RESISC (green).  (LEFT) Model performance for attack strength $\epsilon$. (RIGHT) Model performance for perturbation of the input images as evaluated by the average SSIM of the attacked images. Error bars are not shown, as the standard deviation was on the order of 1$\times10^{-3}$ to 1$\times10^{-6}$. The difference in attack success between target and surrogate models differs from paper Figure \ref{fig:attacks}, indicating an asymmetry between models.}
    \label{fig:sup_attack_eff_res}
\end{figure}

Figure \ref{fig:attacks} of the paper depicts a transfer-based attack with ResNet as the surrogate model and MobileNetv3 as the target model, demonstrating the dependence on attack strength $\epsilon$ and correlating image corruption as measured by the SSIM score are presented for completeness.  In Figure \ref{fig:sup_attack_eff_res}, the symmetric attack with MobileNetv3 acting as the surrogate model and ResNetv2 acting as the target model is shown.  

For the attack strength studies, a curve closer to the upper \textbf{right} corner represents a model which was more robust.  For the image corruption studies, a curve closer to the upper \textbf{left} corner indicates a more robust model.  Solid lines indicate the robustness of the surrogate model to its own attack, and dashed lines indicate the robustness of the target model to transfer-based attack (TBA) generated by the surrogate model.

Figure \ref{fig:sup_attack_eff_res} shows that the choice of surrogate and target model was important, as the performance of a transfer-based attack generated by surrogate model $A$ for target model $B$ does not provide insight into the success of an attack generated with model $B$ as the surrogate and model $A$ as the target.  These results also support a result of this work that the Hausdorff distance between the surrogate embedding and target embedding may vary from the Hausdorff distance when the surrogate model and target model roles are reversed.

\subsection{Persistent Homology as Quantification Approach}

Several instances are presented where the symmetric normalized Hausdorff distance fails to fully capture the similarity (or dissimilarity) between two shared embeddings. This was partly due to the comparative scaling of the data, even after normalization. Persistent homology has emerged as a popular method in Topological Data Analysis (TDA) for evaluating data clustering at different scales; for a comprehensive introduction to the topic, see~\cite{tda}. Persistent homology is useful for understanding latent spaces because it answers the question of what data clusters form at what distances, and what kinds of structures these clusters have.  Persistent homology quantifies how local structure changes for a radius $r=0$ to some threshold distance $r=R$ around each data point. Simplical complexes form and become connected for different values of $r$. The birth $b$ of a connected component is the value of $r$ for which a connected component forms, and the death $d$ of the component is the value of $r$ for which the component is subsumed by another, older component. These birth-death pairs can be used to create a \textit{persistence diagram}~\cite{persistence} as shown in Figure \ref{fig:results_siscore}(d), where $(b,d)$ pairs are plotted on $\mathbb{R}^2$; points closest to the upper left corner $(0, \infty_+)$ are more topologically significant.

The \textbf{bottleneck distance}~\cite{bottleneck} is a metric on the persistence diagram, and may be defined as follows: Let $dgm_1$ and $dgm_2$ be two persistence diagrams. A point cannot die before it was born, restricting all points to the upper left quadrant above the persistence diagram diagonal $\Delta$. Every point in $dgm_1$ was matched to a point in $dgm_2$ and vice versa, and if no match was found, then the nearest point on the persistence diagram diagonal $\Delta$ was chosen~\cite{tda}. The bottleneck distance, $B$, is the minimum distance which allows every birth-death point on the persistence diagram to be mapped to some other point; the distance between all pairwise mappings on the persistence diagram is less than or equal to the bottleneck distance~\cite{tda}. This distance was found by performing a binary search with the Hopcroft-Karp algorithm on the cosine similarity matrix (CSM) of the two birth-death diagrams. Note that there is no guarantee of a unique pair of points for $B$~\cite{tda}. $B \in [0, \infty)$, so that two datasets with identical clustering dynamics and scales will have $B=0$, and a larger bottleneck distance indicates a greater difference in clustering dynamics between two datasets. 

\begin{figure}[!h]
    \centering
    \includegraphics[width=0.8\linewidth]{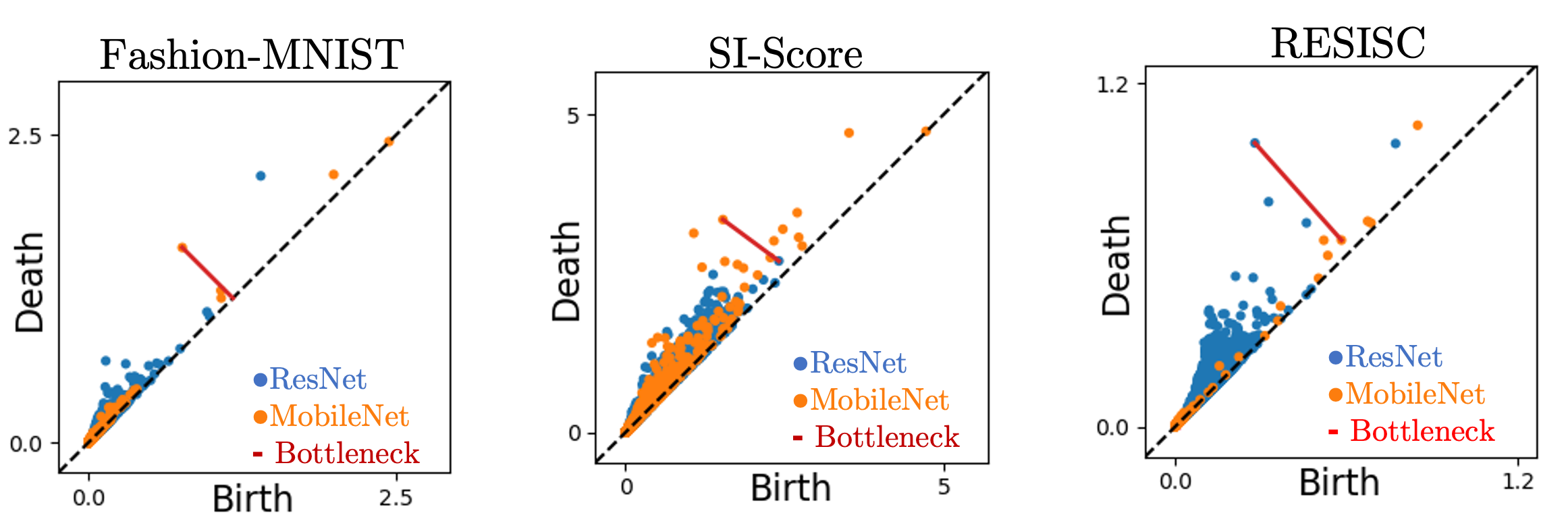}
    \caption{Persistence plots comparing the clustering of MobileNet feature vectors projected with ResNetv2 embeddings onto a shared low dimensional manifold to the clustering of the ResNetv2 embeddings on the same manifold (formally, the $H1$ homology). Each point (blue, orange) on the diagram represents the persistence of a specific $H_1$ connected component in the projection. The bottleneck distance is shown as a red line in each plot.}
    \label{fig:sup_bottleneck}
\end{figure}

Example bottleneck calculations are presented in Figure \ref{fig:sup_bottleneck} for dimensionality reduction using ResNetv2 as the surrogate model and MobileNetv3 as the target model.

In Figure \ref{fig:sup_bottleneck}, the $H_0$ homology was omitted in the persistence diagrams to focus on the Bottleneck analysis relevant to $H_1$. The maximum radius $R$ was chosen to be the Hausdorff distance between the two embeddings (i.e. 1.0), forcing the same scale between UMAP projections and their persistence diagrams. The bottleneck distances are $B_{Fashion-MNIST}=0.41$, $B_{RESISC}= 0.33$, $B_{SI-Score}=0.87$, quantifying the difference in radius required for each pair of embeddings to reach the same connected component structure.

\subsection{Procrustes Analysis as Quantification Approach}

\begin{figure}[!h]
    \centering
    \includegraphics[width=0.7\linewidth]{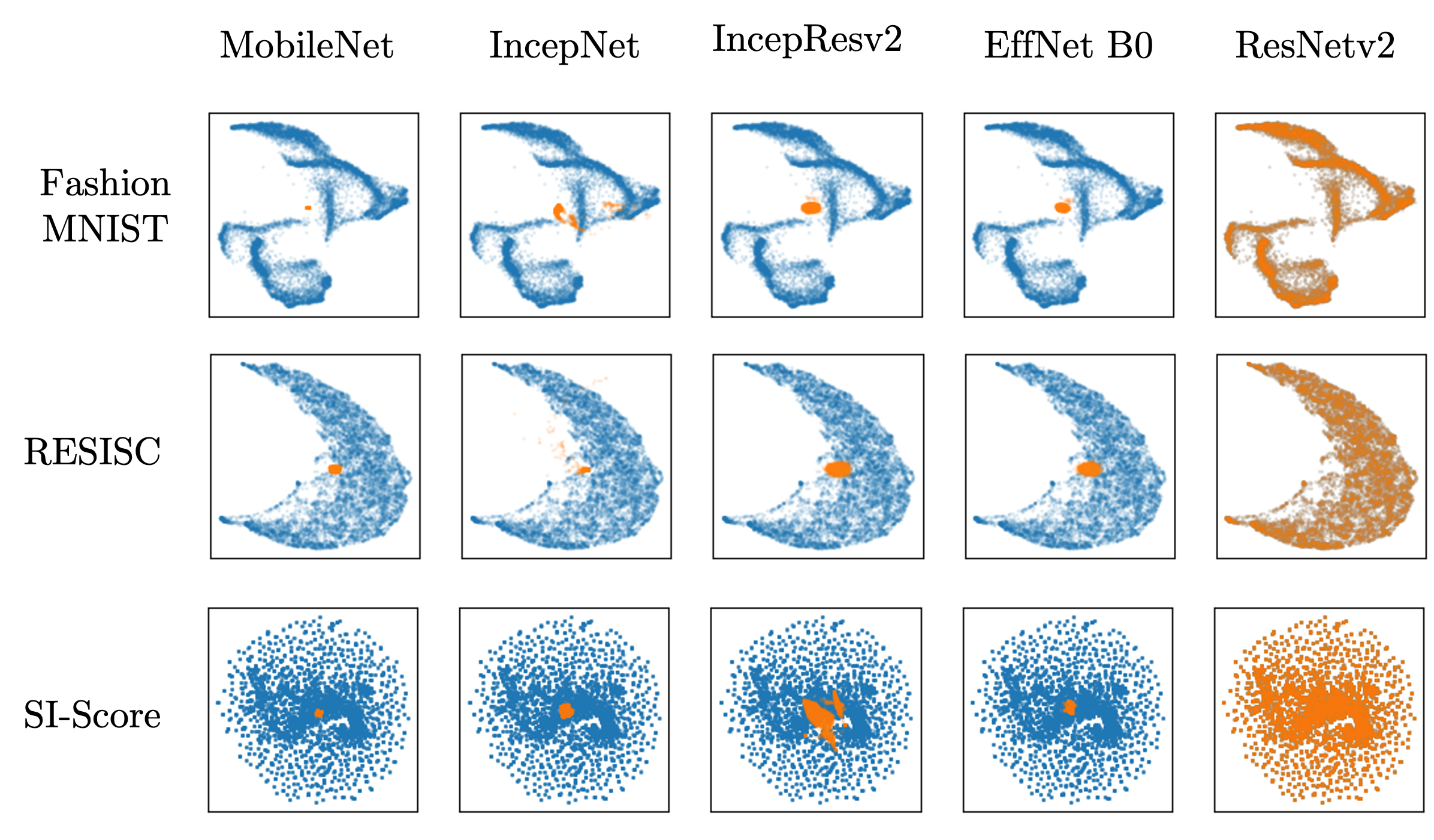}
    \caption{Results of Procrustes analysis comparing projections generated using cross-embedding with the ResNetv2 UMAP embedding of a given dataset.  Projections created from ResNetv2 UMAP embeddings are shown in blue, and while the projection of the feature vectors from another backbone are shown in orange. The Procrustes algorithm does not }
    \label{fig:proc_tform}
\end{figure}

The Procrustes analysis was applied to projections generated exclusively with the ResNetv2 UMAP embeddings.  As shown in Figure \ref{fig:proc_tform}, there seems to be a limit to the Procrustes analysis algorithm's ability to scale the data, with most projected data distributions condensed further into a single point.  Note that this was the case for the MobileNet-FashionMNIST projection, shown in Figure 3 of the paper's main results, which has now been condensed to a single cluster.  

The disparity values $M$ for Figure \ref{fig:proc_tform} are included in Table \ref{tab:disparity_tform}.  For most cases, the $M$ value was 1.00 or very close to 1.00, indicating a maximum disparity.

\begin{table}[!h]
  \caption{Disparity results from Procrustes analysis of projections in Figure \ref{fig:proc_tform}}
  \label{tab:disparity_tform}
  \centering
  \begin{tabular}{lccccccc}
    \toprule
     Backbone & &  FashionMNIST& &  &  RESISC45  & SI-Score  \\
    \midrule
     ResNetv2 &&0.00	&&&0.00	&0.00 \\
    EffNet B0 &&1.00	&&&1.00	&1.00 \\
    IncepResv2 &&1.00	&&&1.00	&1.00 \\
    IncepNet &&1.00	&&&1.00	&0.97 \\
    MobileNet &&1.00	&&&1.00	&1.00 \\
    \bottomrule
  \end{tabular}
\end{table}

The failure of the Procrustes analysis algorithm to maintain structure and scaling of cross-embedded projections motivates the use of the symmetric Hausdorff distance for quantifying changes of cross-embedded projections.

\subsection{Expanded Results for Figure 4}

The results of Figure 4 are presented in tabular form to elucidate the role of different models and datasets.  


\begin{longtable}{ccccc}
    \caption{Numerical results for Figure 4. Some Hausdorff distances were unable to be calculated due to memory constraints; the corresponding attack values are included for completeness. The IncepNetv3 model achieved a maximum F1 score of 0.65 during training, resulting in potentially suppressed AA values; these values are denoted by the * symbol.}
    \label{tab:raw_results} \\

    \toprule
    Target & Surrogate & Dataset & AA$(\epsilon = 0.03)$ & Hausdorff Distance $H$ \\
    \midrule
    \endfirsthead

    \caption[]{(Continued)}\\
    \toprule
    Target & Surrogate & Dataset & AA$(\epsilon = 0.03)$ & Hausdorff Distance $H$ \\
    \midrule
    \endhead

    \midrule
    \multicolumn{5}{r}{\textit{Continued on next page}} \\
    \endfoot

    \bottomrule
    \endlastfoot
    
    EfficientNet B0 & IncepNet v3 & Fashion-MNIST & 0.61 & 0.88 \\
    EfficientNet B0 & IncepNet v3 & RESISC & 0.79 & - \\
    EfficientNet B0 & IncepNet v3 & SISCORE & 0.97 & 0.4 \\
    EfficientNet B0 & IncepNetRes & Fashion-MNIST & 0.70 & 0.52 \\
    EfficientNet B0 & IncepNetRes & RESISC & 0.79 & - \\
    EfficientNet B0 & IncepNetRes & SISCORE & 0.96 & - \\
    EfficientNet B0 & MobileNet & Fashion-MNIST & 0.72 & 0.96 \\
    EfficientNet B0 & MobileNet & RESISC & 0.74 & 0.59 \\
    EfficientNet B0 & MobileNet & SISCORE & 0.95 & 0.55 \\
    EfficientNet B0 & ResNet & Fashion-MNIST & 0.62 & 0.81 \\
    EfficientNet B0 & ResNet & RESISC & 0.78 & 0.73 \\
    EfficientNet B0 & ResNet & SISCORE & 0.96 & 0.27 \\
    IncepNet v3 & EffNet & Fashion-MNIST & 0.10 & 0.60 \\
    IncepNet v3 & EffNet & RESISC & 0.60* & 0.42 \\
    IncepNet v3 & EffNet & SISCORE & 0.79 & 0.21 \\
    IncepNet v3 & IncepNetRes & Fashion-MNIST & 0.10 & 0.72 \\
    IncepNet v3 & IncepNetRes & RESISC & 0.58* & 0.49 \\
    IncepNet v3 & IncepNetRes & SISCORE & 0.82 & - \\
    IncepNet v3 & MobileNet & Fashion-MNIST & 0.10 & 0.96 \\
    IncepNet v3 & MobileNet & RESISC & 0.55* & 0.29 \\
    IncepNet v3 & MobileNet & SISCORE & 0.83 & 0.23 \\
    IncepNet v3 & ResNet & Fashion-MNIST & 0.10 & 0.81 \\
    IncepNet v3 & ResNet & RESISC & 0.59* & 0.73 \\
    IncepNet v3 & ResNet & SISCORE & 0.85 & 0.21 \\
    IncepNetRes & EffNet & Fashion-MNIST & 0.06 & 0.58 \\
    IncepNetRes & EffNet & RESISC & 0.79 & 0.49 \\
    IncepNetRes & EffNet & SISCORE & 0.91 & 0.24 \\
    IncepNetRes & IncepNet v3 & Fashion-MNIST & 0.06 & 0.87 \\
    IncepNetRes & IncepNet v3 & RESISC & 0.73 & 0.48 \\
    IncepNetRes & IncepNet v3 & SISCORE & 0.94 & 0.35 \\
    IncepNetRes & MobileNet & Fashion-MNIST & 0.10 & 0.96 \\
    IncepNetRes & MobileNet & RESISC & 0.82 & 0.30 \\
    IncepNetRes & MobileNet & SISCORE & 0.93 & 0.33 \\
    IncepNetRes & ResNet & Fashion-MNIST & 0.07 & 0.81 \\
    IncepNetRes & ResNet & RESISC & 0.74 & 0.73 \\
    IncepNetRes & ResNet & SISCORE & 0.96 & 0.21 \\
    MobileNet & EffNet & Fashion-MNIST & 0.19 & 0.36 \\
    MobileNet & EffNet & RESISC & 0.70 & 0.20 \\
    MobileNet & EffNet & SISCORE & 0.97 & 0.21 \\
    MobileNet & IncepNet v3 & Fashion-MNIST & 0.23 & 0.46 \\
    MobileNet & IncepNet v3 & RESISC & 0.70 & - \\
    MobileNet & IncepNet v3 & SISCORE & 0.92 & 0.21 \\
    MobileNet & IncepNetRes & Fashion-MNIST & 0.21 & 0.28 \\
    MobileNet & IncepNetRes & RESISC & 0.72 & - \\
    MobileNet & IncepNetRes & SISCORE & 0.89 & - \\
    MobileNet & ResNet & Fashion-MNIST & 0.79 & 0.12 \\
    MobileNet & ResNet & RESISC & 0.73 & 0.22 \\
    MobileNet & ResNet & SISCORE & 0.88 & 0.09 \\
    ResNet & EffNet & Fashion-MNIST & 0.67 & 0.57 \\
    ResNet & EffNet & RESISC & 0.75 & 0.42 \\
    ResNet & EffNet & SISCORE & 0.97 & 0.42 \\
    ResNet & IncepNet v3 & Fashion-MNIST & 0.81 & 0.71 \\
    ResNet & IncepNet v3 & RESISC & 0.90 & 0.48 \\
    ResNet & IncepNet v3 & SISCORE & 0.97 & 0.22 \\
    ResNet & IncepNetRes & Fashion-MNIST & 0.81 & 0.81 \\
    ResNet & IncepNetRes & RESISC & 0.75 & 0.43 \\
    ResNet & IncepNetRes & SISCORE & 0.96 & 0.21 \\
    ResNet & MobileNet & Fashion-MNIST & 0.25 & 0.91 \\
    ResNet & MobileNet & RESISC & 0.73 & 0.39 \\
    ResNet & MobileNet & SISCORE & 0.97 & 0.28 \\
\end{longtable}

\end{document}